\documentclass[letterpaper, 10 pt, conference]{ieeeconf}
\IEEEoverridecommandlockouts    
\usepackage{graphics}           
\usepackage{times}              
\usepackage{amsmath}            
\usepackage{amssymb}            
\usepackage{graphicx}
\usepackage{algorithm}
\usepackage[noend]{algpseudocode}
\usepackage{booktabs}
\usepackage{color}
\usepackage[nocompress]{cite} 
\definecolor{instructioncolor}{rgb}{.5,.5,.5}

\usepackage[font=small]{caption}

\def\secref#1{Sec.~\ref{#1}}
\def\figref#1{Fig.~\ref{#1}}
\def\tabref#1{Tab.~\ref{#1}}
\def\eqref#1{Eq.~(\ref{#1})}
\def\algref#1{Alg.~\ref{#1}}

\makeatletter
\usepackage{xspace}
\DeclareRobustCommand\onedot{\futurelet\@let@token\@onedot}
\def\@onedot{\ifx\@let@token.\else.\null\fi\xspace}

\def\etal{{et al}\onedot}
\makeatother

\usepackage{array}
\newcolumntype{L}[1]{>{\raggedright\let\newline\\\arraybackslash\hspace{0pt}}m{#1}}
\newcolumntype{C}[1]{>{\centering\let\newline\\\arraybackslash\hspace{0pt}}m{#1}}
\newcolumntype{R}[1]{>{\raggedleft\let\newline\\\arraybackslash\hspace{0pt}}m{#1}}

\newcommand{\bigO}[1]{$\mathcal{O}(#1)$}

\newcommand{\norm}[1]{\lVert#1\lVert}

\renewcommand{\b}[1]{\mbox{\boldmath$#1$}}

\newcommand{\bn}{\b n}

\newcommand{\bp}{\b p}

\newcommand{\bv}{\b v}

\newcommand{\linelabels}{\textit{L}^{\mathrm{lines}}}
\newcommand{\linelabelsrle}{\textit{L}^{\mathrm{rle}}}
\newcommand{\pointlabels}{\textit{L}^{\mathrm{points}}}

\title{\LARGE \bf Fast and Robust Normal Estimation for Sparse {LiDAR} Scans}

\author{Igor Bogoslavskyi \and Konstantinos Zampogiannis \and Raymond Phan%
  \thanks{All authors are with Magic Leap.}%
}

\begin{document}
\maketitle
\thispagestyle{empty}
\pagestyle{empty}

\begin{abstract}
Light Detection and Ranging (LiDAR) technology has proven to be an important part of many robotics systems. Surface normals estimated from LiDAR data are commonly used for a variety of tasks in such systems. As most of the today's mechanical LiDAR sensors produce sparse data, estimating normals from a single scan in a robust manner poses difficulties.

In this paper, we address the problem of estimating normals for sparse LiDAR data avoiding the typical issues of smoothing out the normals in high curvature areas.

Mechanical LiDARs rotate a set of rigidly mounted lasers. One firing of such a set of lasers produces an array of points where each point's neighbor is known due to the known firing pattern of the scanner. We use this knowledge to connect these points to their neighbors and label them using the angles of the lines connecting them. When estimating normals at these points, we only consider points with the same label as neighbors. This allows us to avoid estimating normals in high curvature areas.

We evaluate our approach on various data, both self-recorded and publicly available, acquired using various sparse LiDAR sensors. We show that using our method for normal estimation leads to normals that are more robust in areas with high curvature which leads to maps of higher quality. We also show that our method only incurs a constant factor runtime overhead with respect to a lightweight baseline normal estimation procedure and is therefore suited for operation in computationally demanding environments.
\end{abstract}

\section{Introduction}
\label{sec:intro}

LiDAR sensors have proven their versatility in a number of domains, ranging from self-driving cars~\cite{marcuzzi2023ral,nunes2022ral-3duis,chen2021icra,vizzo2021icra}, through agriculture~\cite{chebrolu2020icra,riviera2023cea}, and even on space missions~\cite{pedersen2012isairas}. In all of these environments, LiDAR sensors are routinely used for perception, localization and mapping among other use cases. Even though our contributions might be useful in all of these scenarios, in this paper we focus on applying our normal estimation method to the latter ones: localization and mapping tasks using LiDAR data.
\begin{figure}[t]
  \centering
  \includegraphics[width=0.95\linewidth]{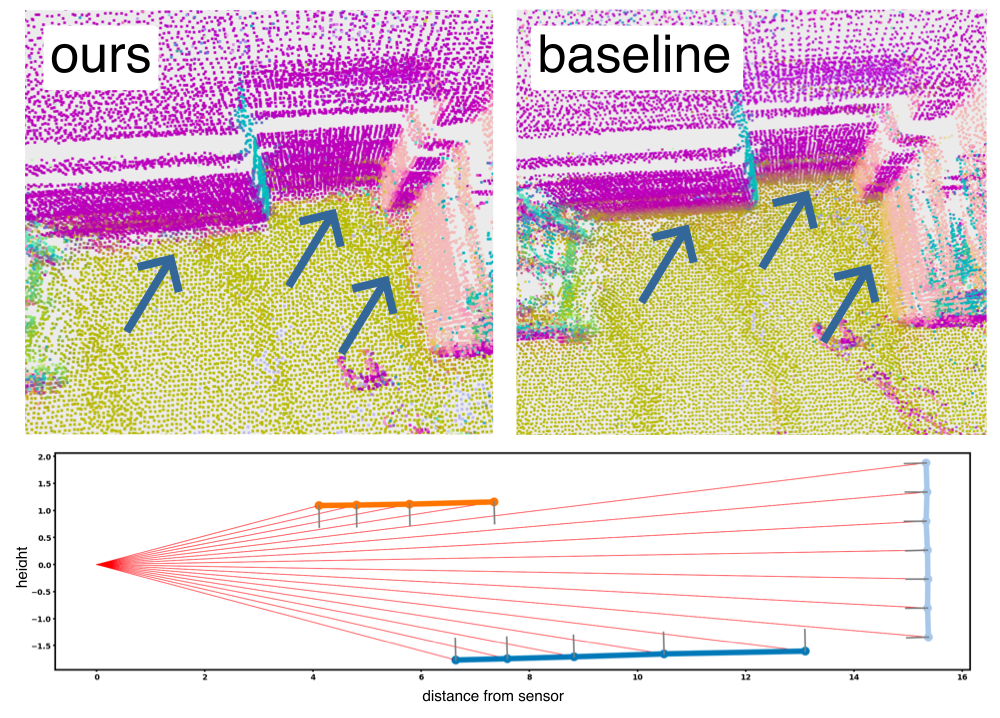}
  \caption{Top two images show a part of an environment reconstructed by aligning multiple individual point clouds from the HILTI dataset~\cite{hilti}. Both point clouds are colored by the normals of their points. The right one uses a baseline implementation of normals and arrows indicate areas where the normals spill into an orthogonal surface. The left image shows that this does not happen when using the proposed method for normal estimation. The bottom image shows a single scan generated from a Velodyne VLP16 LiDAR, positioned at $(0, 0)$, with gray lines showing the normals estimated with the proposed method. Thick colored lines connecting the points indicate the connected components that the points belong to. }
  \label{fig:motivation}
\end{figure}

A key step in these tasks is 3D data alignment and a key choice for data alignment is the choice of features. There is a significant amount of research into which features are best for the task. Today's SLAM systems use corner and planar features~\cite{zhang2014rss}, surfels~\cite{behley2018rss}, or even raw points directly~\cite{vizzo2023ral}. However, we believe that a point-to-plane metric remains an important staple in point cloud data alignment. In our view, it allows for better data associations, which simplifies and improves the alignment process, warranting the cost of additional computations needed for normal feature extraction.

There is a number of methods to compute normals needed for the point-to-plane metric to function. One common method is to fit a plane to a neighborhood of each point~\cite{hoppe1992siggraph} and use the normal of that plane as an estimate of the surface orientation in that point. This method, however, has a number of drawbacks: it works poorly on sparse data, where larger neighborhoods must be used, and, being of \bigO{\log n} complexity, it takes a significant amount of computation to find the neighborhood of any given point.

Therefore, to reduce the amount of computation needed, we follow the approach used by Behley and Stachniss~\cite{behley2018rss}, which we will call the \emph{baseline} method for normal computation. They utilize the fact that the data produced by a LiDAR sensor is structured due to the LiDAR data acquisition pattern. Mechanical LiDARs spin an array of rigidly fixed lasers, producing data that can be viewed in an organized fashion. This allows to query the neighbors of any point in \bigO{1} time which makes the normal computation very efficient. This method is faster than the full plane-fitting method but suffers from the fact that the data can be sparse, especially between the points stemming from different lasers, due to the physical laser diode placement in the body of a LiDAR sensor. Thus baseline normal estimation fails to produce satisfying results when the curvature of the underlying surface is high.

The main contribution of this paper is a method that improves upon the baseline normal computation technique by clustering the points stemming from neighboring lasers into components likely describing the same underlying surface and computing the normals within the clusters of these points, avoiding cross-surface normal computation. We evaluate our method with respect to our main two claims: (i) that our method produces normals that are more robust than the ones estimated by the baseline method and (ii) that our method only incurs a constant factor runtime overhead when compared to the baseline method.

\section{Related Work}
\label{sec:related}

The topic of normal estimation enjoyed an extensive coverage throughout the years. As of this
writing, methods for estimating surface normals from point clouds can be decomposed into two
predominant methods: \emph{traditional} methods and \emph{learning-based} methods.

The traditional methods use the geometry and/or extracting hand-crafted features from the point cloud directly to find the best fitting plane for a subsequent a least-squares minimization. Our method also lies within this category.

One of the earliest methods for estimating surface normals in point clouds was to use PCA~\cite{hoppe1992siggraph} which derives each point's normal by computing the eigenvalues and eigenvectors of the covariance matrix with the $k$-nearest neighbors. The eigenvector corresponding to the smallest eigenvalue is determined to be the normal at that point.
Stemming from this method, several other approaches have been proposed~\cite{avron2010acmtransgraph,mitra2003scg,yoon2007cad} which consider noise, curvature and sampling density of the point clouds. However, these methods tend to smooth out sharp edges and corners in potential shapes that are apparent in the point cloud which is what we are trying to avoid with our method.

To combat these problems, other approaches like Voronoi-based methods~\cite{alliez2007sgp,amenta98scg,dey2005cgf}, minimizing $L_0$ or $L_1$ norms~\cite{avron2010acmtransgraph,sun2015cag}, adopting statistical-based methods~\cite{li2010cag,yu2019ame} and making use of low-rank matrices~\cite{zhang2013cag,liu2015cag,lu2022tvcg} have been proposed. These methods have
a better ability to preserve sharp features for point clouds in the presence of noise. Similarly, in the work that is closest to ours, Badino~\etal~\cite{badino2011icra} reformulate the estimation problem by taking advantage of the organized nature of the data available from a LiDAR scanner, and use range images to compute normals. The normals then are obtained by calculating derivatives from the surface generated from a spherical range image. All of the methods above, though, work with denser data, while we explicitly target to improve normals computed from a single revolution of a very sparse LiDAR.

One typical way to estimate normals in a computationally-constrained environment is to perform a cross product between the vectors towards the points neighboring the query point. Behley and Stachniss~\cite{behley2018rss} use the structured nature of the data provided by a LiDAR sensor to get the neighbors of such a point in constant time and compute a normal using the cross product as described above. Our method builds directly on top of this idea improving it by labeling the points in correspondence to the underlying surface in the environment that they stem from.

In recent years, learning-based normal estimation methods have become more prevalent with the proliferation of deep learning. Only a few of these methods have the ability to preserve sharp features, robustly handle noise and generalize across different shapes. Most of these approaches still require a GPU to run fast enough to be used effectively. We do not aim to compete with these methods and provide them here for a general overview.

One of the first approaches was presented by Boulch and Marlet~\cite{boulch2016cgf} where the HoughCNN architecture was introduced. In it 3D point clouds are mapped to a Hough space, and a CNN estimates normals in this representation. The transformation of the 3D points into a 2D Hough space is required as CNNs operate on 2D or image data. Because of this transformation, this method may discard vital geometrical details. Alternatively, in their paper, Qi~\etal~\cite{qi2017cvpr} introduce their seminal PointNet architecture that allows neural networks to directly consume 3D point clouds without transforming them into a 2D space. Using PointNet as a backbone, Guerrero~\etal~\cite{guerrero2018cgf} introduce their PCPNet for feature extraction to encode local neighborhoods of points to allow for direct regression of the normals. Similarly, Ben-Shabat~\etal~\cite{benshabat2019cvpr} propose a bagging approach that uses multiple networks in a mixture that observes the neighborhoods of each point at varying scales and selects the optimal network for predicting the normals. In \cite{hyeon2019ars}, the authors propose a multi-scale $k$-nearest neighbors to robustly extract local features used as a pre-processing step to finally estimate the normals. Most of these and similar learning-based works~\cite{hyeon2019ars,wang2020bmvc, lenssen2020bmvc,wang2022icme,li2022neurips} aim at finding the normals of dense and unstructured point clouds and due to the need of a GPU do not warrant a direct comparison to our method which aims to run fast on any CPU.

To the degree of our knowledge, our method is the first method to perform connected components on structured data coming from a sparse LiDAR sensor to aid the normal robustness while still maintaining the constant normal computation time per point.

\section{Our Approach to Fast and Robust Normal Computation}
\label{sec:main}

\subsection{Baseline Normal Computation}
\label{sub:baseline}

\begin{figure}[t]
  \centering
  \includegraphics[width=0.6\linewidth]{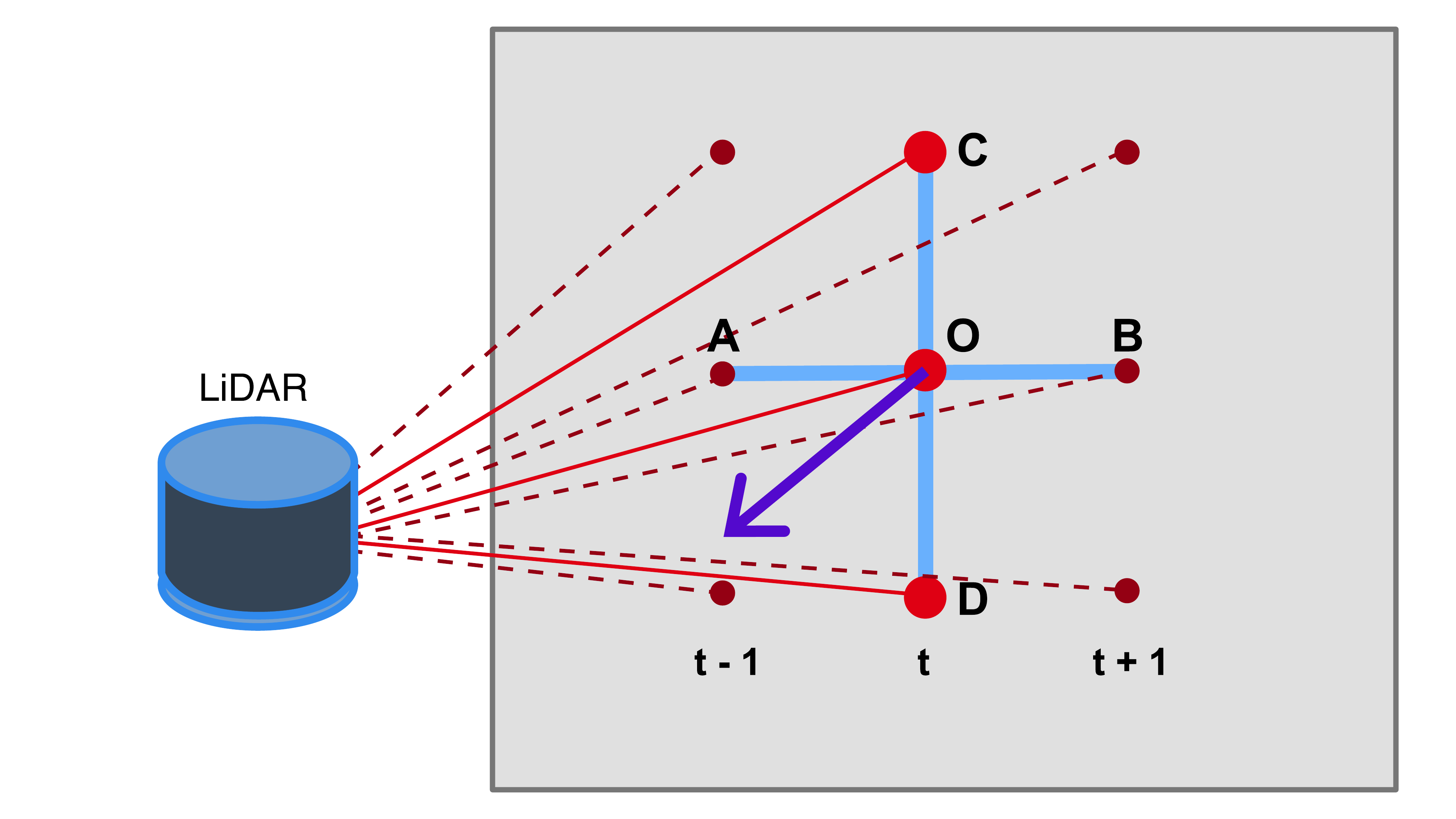}
  \caption{Data that comes from a mechanical LiDAR is organized. Point $O$ has 4 neighbors. Points $A$ and $B$ are generated by the same beam at a previous and next scan. Points $C$ and $D$ are generated by the beams neighboring the beam from which the point $O$ stems within the same scan. Given these 4 neighbors a baseline way to compute a normal would be to compute a cross product between vectors $\overrightarrow{AB}$ and $\overrightarrow{CD}$ as shown in~\eqref{eq:normals} that results in the normal shown in purple. This approach is noisy but serves as a good starting point for our approach.}
  \label{fig:normals_naive}
\end{figure}

Our approach improves on the baseline normal estimation method as presented by Behley and Stachniss~\cite{behley2018rss}. Likewise to their work, we focus on data generated by mechanical LiDARs. These LiDARs spin a physical array of lasers and provide the returns of these lasers as they are measured. These data can be viewed in an organized manner as illustrated in~\figref{fig:normals_naive}, where the shown points stem from lasers physically mounted on top of each other. Such data storage allows us to query the immediate neighbors of any given point in constant time. Consider that every point $\bp = [x, y, z]^\top$ has up to four neighbors: $\bp_\mathrm{left}$, $\bp_\mathrm{right}$, $\bp_\mathrm{top}$, and $\bp_\mathrm{bottom}$. Given these neighbors we can compute the normal $\bn = [n_x, n_y, n_z]^\top$ at point $\bp$ as follows~\cite{behley2018rss}:

\begin{equation}
  \label{eq:normals}
  \bn = (\bp_\mathrm{right} - \bp_\mathrm{left}) \times (\bp_\mathrm{top} - \bp_\mathrm{bottom}),
\end{equation}
here, $(\bp_\mathrm{right} - \bp_\mathrm{left})$ corresponds to $\overrightarrow{AB}$ in~\figref{fig:normals_naive}, while $(\bp_\mathrm{top} - \bp_\mathrm{bottom})$ corresponds to $\overrightarrow{CD}$, and the normal $\bn$ being shown in purple. One thing to consider in this approach is that if one of either horizontal or vertical neighbors is missing, the point $\bp$ can be used instead. Should both neighbors in one direction be missing, we consider such a point invalid and do not compute a normal for it.

This method has its advantages and disadvantages. Because the neighborhood of any single point is found in~\bigO{1} time, it is efficient and thus is commonly used when the computational resources are limited. At the same time, such small neighborhoods are a double-edged sword as this method is sensitive to the noise in the positions of the points. While white noise induced by the sensor noise does not usually pose big issues as the normal vectors can be averaged over multiple measurements, there is also a \emph{systematic error caused by the sparsity of the LiDAR data}, especially in the data recorded by neighboring lasers which leads to points far apart being wrongly considered as neighbors.

\begin{figure}[t]
  \centering
  \includegraphics[width=0.5\linewidth]{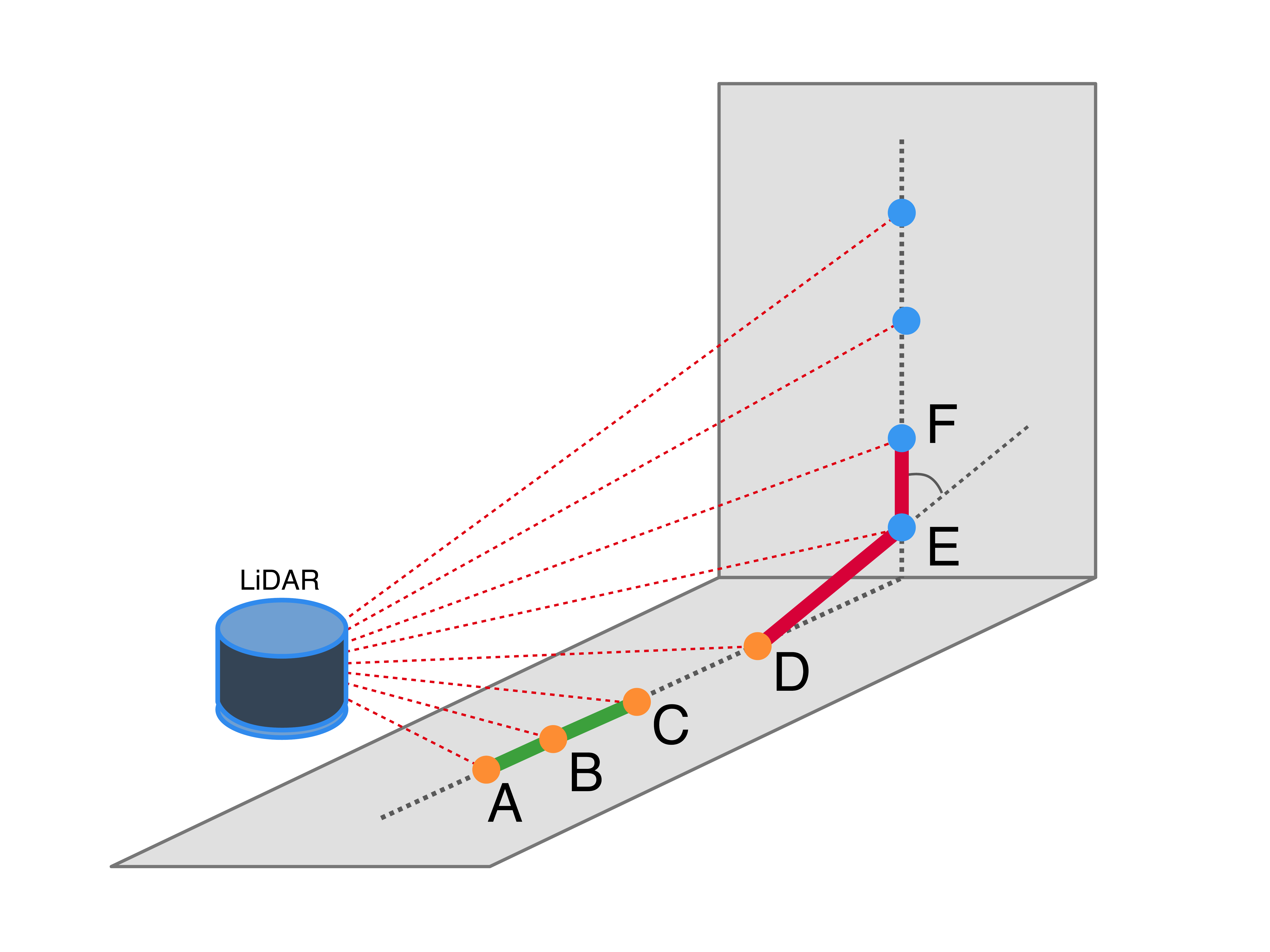}
  \caption{Points generated from a single vertical stack of lasers. We interchangeably call such points a slice of LiDAR data or being "vertical neighbors". Points are colored orange and blue with respect to their label after clustering them with respect to the angles of the line segments connecting them. Some of these segments are shown too. The green segments $AB$ and $BC$ which form an angle $\angle ABC$ indicate that the points should belong to the same connected component, while the red lines forming the angle $\angle DEF$ indicate that the points $D$, $E$, and $F$ should belong to different connected components.}
  \label{fig:normals}
\end{figure}

The underlying assumption here is that the points lie close to each other and therefore stem from the same underlying surface patch. The distance between the vertical neighbors, however, grows quickly when the points are measured further away from the sensor. This reduces the likelihood that the neighboring points stem from the same underlying surface, breaking the assumption above. For an example of such a situation, see~\figref{fig:normals}, where the points $D$ and $E$ lie on different planar patches of the environment while still being neighbors due to their acquisition order. Our main contribution is a heuristic-based method that deals with such situations better, increasing the chances of estimating normals within a single planar patch.

It is important to note that such data organization is an approximation due to the movement of the LiDAR sensor through the environment during data acquisition. The effect of this is typically low and various motion compensation techniques can be applied in order to mitigate this effect and we therefore neglect it in the remainder of the paper.

\subsection{Create Run Length Encoding of Line Labels}
\label{sub:linelabels}

In our previous work on LiDAR point cloud clustering~\cite{bogoslavskyi2016iros,bogoslavskyi2017pfg} we segmented LiDAR data using connected components analysis in order to detect a ground plane. In that work, we analyzed the line segments formed by connecting the vertical neighbors within the point cloud stemming from a single revolution of a mechanical LiDAR. Should the difference in the inclination of such line segments be small, the next point is labeled as ground.

In this work we modify and extend that idea while using a similar approach to separate points that stem from different planar segments in the underlying data. Ideally for a set of points stemming from a vertical array of lasers shown in~\figref{fig:normals}, we would like to make sure that points $A$, $B$ and $C$ end up being in the same connected component, while the points $D$, $E$ and $F$ are split into separate components resulting in the final point labels to be as indicated by the colors of the points in the figure.

In this section, we focus on separating the line segments formed by the points within a single vertical LiDAR scan (similar to points in~\figref{fig:normals}) into connected components based on the differences in the line segment orientation. Because we only consider a single stack of lasers here we can talk about such points in 2D. The same logic can, of course, be applied in the horizontal direction without the loss of generality. That being said, it is less important to apply our approach in the horizontal direction as the distances between the points generated by the same laser diode are much smaller than the distances between the points generated by neighboring laser diodes, making it less likely for the points to land on different surfaces, thus making the usage of a more complex method harder to motivate.

\begin{algorithm}[t]
  {\footnotesize

    \caption{Point Labeling}
    \label{alg:clustering}
    \begin{algorithmic}[1]
      \Function{LabelPoints}{\textit{P}, $\alpha_{\mathrm{threshold}}$}
      \State $\linelabelsrle \gets \texttt{EncodeLines}(\textit{P}, \alpha_{\mathrm{threshold}})$
      \State $\pointlabels \gets \texttt{ExpandLabels}(\textit{P}, \linelabelsrle)$
      \State \Return~$\pointlabels$
      \EndFunction{}

      \vspace{1mm}

      \Function{EncodeLines}{\textit{P}, $\alpha_{\mathrm{threshold}}$}
      \State$\bv_\mathrm{prev} \gets \textit{P}_1 - \textit{P}_0$
      \State$\linelabelsrle \gets \{\}$
      \State$c \gets 1$
      \For{$i = 1 \ldots \norm{\textit{P}}$}
      \State$\bv_\mathrm{next} \gets \textit{P}_{i + 1} - \textit{P}_i$
      \State $\alpha \gets \angle(\bv_\mathrm{prev}, \bv_\mathrm{next})$ using~\eqref{eq:angle}.
      \If{$\alpha > \alpha_{\mathrm{threshold}}$}
      \State $\linelabelsrle \gets \linelabelsrle \cup \{c\}$, $c \gets 0$;
      \EndIf{}
      \State $c \gets c + 1$
      \State$\bv_\mathrm{prev} \gets \bv_\mathrm{next}$
      \EndFor{}
      \State $\linelabelsrle \gets \linelabelsrle \cup \{c\}$
      \State \Return~$\linelabelsrle$
      \EndFunction{}

      \vspace{1mm}

      \Function{ExpandLabels}{$\textit{P}, \linelabelsrle$}
      \State Initialize point labels: $\pointlabels \gets \{0\}$
      \State Previous component strength: $s_{\mathrm{prev}} = 1$
      \For {$l = 0 \ldots \norm{\linelabelsrle}$}
      \State Current component strength: $s \gets \linelabelsrle_l$
      \State Index of disputed point: $i_d \gets \norm{\pointlabels} - 1$
      \If{$s > 1$ \textbf{and} $s_{\mathrm{prev}} == 1$}
      \State $\pointlabels_{i_d} \gets l$
      \ElsIf{$s > 1$ \textbf{and} $s_{\mathrm{prev}} > 1$}
      \If{$\norm{\textit{P}_{i_d} - \textit{P}_{i_d + 1}} < \norm{\textit{P}_{i_d} - \textit{P}_{i_d - 1}}$}
      \State $\pointlabels_{i_d} \gets l$
      \EndIf
      \EndIf{}
      \State $\pointlabels \gets \pointlabels \cup \{l\}^s$
      \State $s_{\mathrm{prev}} \gets s$
      \EndFor{}
      \State \Return~$\pointlabels$
      \EndFunction{}
    \end{algorithmic}
  }
\end{algorithm}

The approach we use for finding connected components can be seen as a trivial case of the depth first search (DFS) on a graph that degenerates into a chain. The nodes of such a graph are line segments between the neighboring points and the weights on the edges represent differences in the line segment orientation computed by the following equation:

\begin{equation}
  \label{eq:angle}
  \alpha = \arccos \frac{
    \bv_1 \cdot \bv_2
  }{
    \norm{\bv_1} \norm{\bv_2}
  }.
\end{equation}

The connected component algorithm walks over a sequence of lines and compares if the angle difference between them is larger than a given threshold. If the angle difference is below the threshold, the next line is labeled with the same label as the previous one, otherwise it gets the next consecutive label.

To store the line labels along with the size of their component, that we call "strength", we make use of run length encoding (RLE)~\cite{runlength}. Let us assume for the sake of example that while performing the connected component labeling we ended up seeing such a string of line labels:
\begin{equation}
  \linelabels = \{0, 0, 0, 0, 1, 2, 2, 2, 3, 4, 5, 5, 5\}.
\end{equation}
This list can be compactly represented with the RLE as a sequence of entries of the form $\{^n l\}$, where $n$ is the number of times a label $l$ appears in the label list. For an example shown above the RLE will look as follows:
\begin{equation}
  \linelabelsrle = \{^4 0,^1 1,^3 2,^1 3,^1 4,^3 5\}.
\end{equation}
Utilizing the fact that our labels are consecutive integer numbers, we can further simplify the representation of such a list as an array of numbers indicating the "strength" of the component with index $l: [4, 1, 3, 1, 1, 3]$. Not only is this representation more compact than the full list of labels, it also makes the next step of labeling points easier.

Function \texttt{EncodeLines} in~\algref{alg:clustering} shows this algorithm in detail. Here, \textit{P} represents a single vertical stack of LiDAR points, with $\textit{P}_i$ representing the $i$-th point in it, $\alpha_{\mathrm{threshold}}$ is the angle difference threshold, and $c$ indicates the current component strength.

\subsection{Point Labeling Using RLE}
\label{sub:pointlabels}

To compute the normals that do not cross the border of the underlying planar patches, we need to extend the labels of the line segments onto the points. We do this by walking over the $\linelabelsrle$ values and filling the $\pointlabels$ as described in the function \texttt{ExpandLabels} in~\algref{alg:clustering}. The main idea of such label expansion is that we want to assign points that are adjacent to a "strong" component the label of that component. We call a component "strong" if its strength recorded in the $\linelabelsrle$ is bigger than 1. Points that find themselves neighboring on two strong components get assigned to the one closest to them in Euclidean space. Points stuck between two weak components are considered to have high curvature. These points get an arbitrary label of one of their adjacent line labels and can be handled down the line. We can choose not to compute a normal in such points or mark them as points with high curvature and handle them later. The underlying assumption behind this method is that the points that form a "strong" components lie on smooth surfaces.

\subsection{Normal Computation using Point Labels}
\label{sub:normalcomp}
Once the points are labeled, computing their normals follows the baseline approach described in~\secref{sub:baseline} closely. We compute the normal with~\eqref{eq:normals} using the cross product between the vectors spanning between the neighboring points. The main difference to the baseline approach, however, is that for a point with label $l$ we only consider neighbors that have the same label $l$. This results in points that lie on the border between two labels to only consider their one neighbor that has the same label as themselves, improving the normal estimation in areas with high curvature. One corner case in such an approach appears when a point has both of its vertical neighbors having different label from itself. In such a case, we argue that no reliable normal can be computed and avoid normal computation of such a point in our experiments. That being said, these points can instead be marked as points exhibiting high curvature and dealt with at a later point in the pipeline.

The overall approach is shown in~\algref{alg:clustering} but can be optimized when being efficiently implemented. While we show a two-step approach to aid explanation, further optimizations can be made that reduce the approach to a single pass over the data, yielding minimal overhead with respect to the baseline approach.

Our approach is a heuristic-based approach that relies on an assumption that an environment consists of multiple smooth components which might not be the case at all times. However, as we show in our experimental evaluation, our approach tends to work well in practice in a variety of scenarios.

\section{Experimental Evaluation}
\label{sec:exp}

%
The main focus of this work is a system that detects normals on sparse data produced by a single LiDAR. It is able to avoid a typical pitfall of baseline implementations by clustering the points that stem from a single vertical array of lasers and only considering the points with the same label for normal estimation.

%
To this end, the results of our experiments support our key claims, which are:
(i) The proposed method for normal computation produces more robust normals in the areas of high curvature;
(ii) The normals computation only incurs a linear runtime overhead with respect to the baseline method;

\subsection{Experimental Setup}
\label{sub:expsetup}

Before we dive into the experiments, we would like to outline a SLAM system used in some of these experiments. While this SLAM system is not part of the contributions of this work, we believe it is helpful to understand how it works to interpret the presented results better. We employ this system in all experiments that support our first claim of our method producing normals with better robustness in the high curvature parts of the environment.

\begin{figure}[t]
  \centering
  \includegraphics[width=0.99\linewidth]{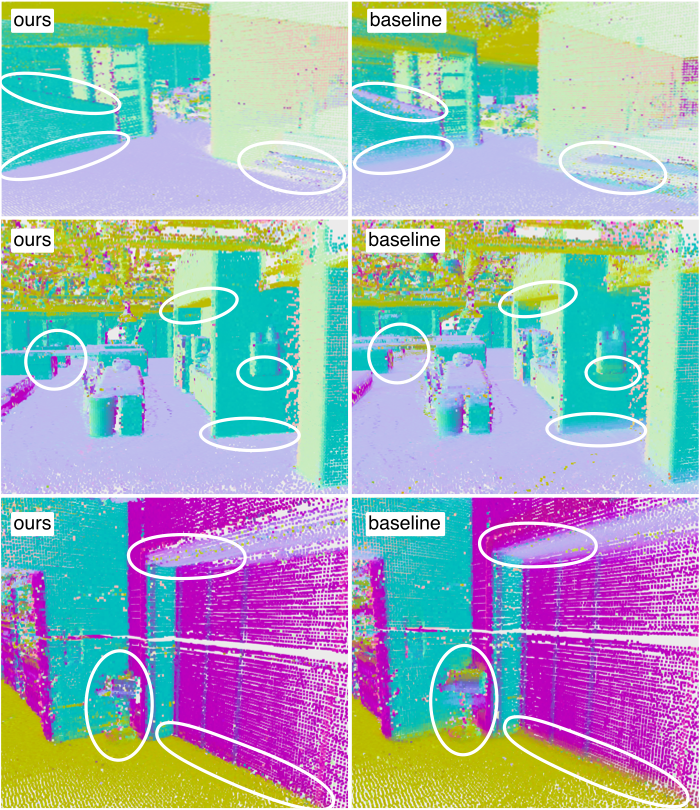}
  \caption{A comparison of the robustness of the normals on various datasets. All of the images show point clouds as aggregated by our SLAM system described in~\secref{sub:expsetup} with the colors representing the normals. Every row of images shows the map with our normals on the left and a map with baseline normals on the right. The top two rows are recorded with a single Velodyne VLP16 LiDAR through a NavVis M6 rig in our office, while the bottom row is reconstructed from the \texttt{BASEMENT\_3} sequence of the HILTI 2021 dataset~\cite{hilti} recorded with a single Ouster OS0 64 beam LiDAR that we subsample to 32 beams to simulate sparsity. The white ellipses highlight parts of the environment that are of interest and exhibit a substantial change in normal robustness. Note how our approach consistently provides sharper gradient in normals.}
  \label{fig:robustness}
\end{figure}

\begin{figure*}[t]
  \centering
  \includegraphics[width=0.35\linewidth]{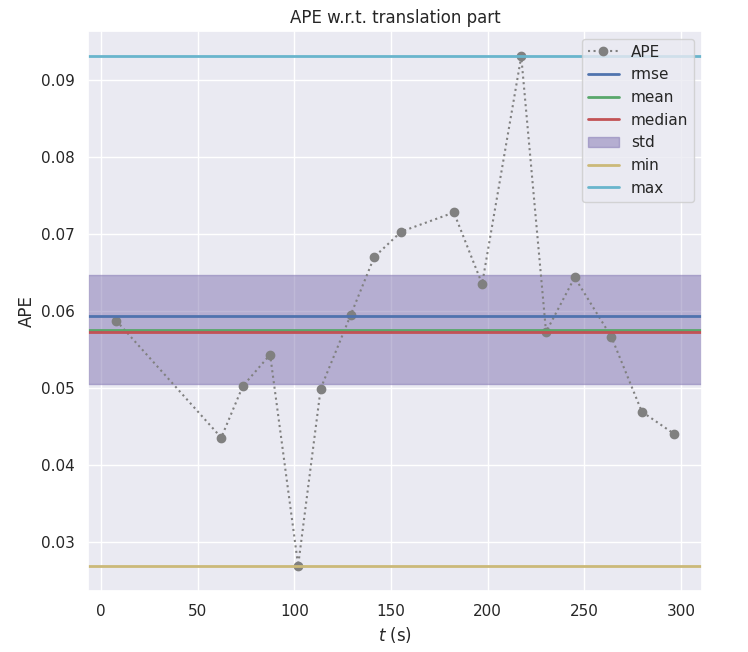}
  \includegraphics[width=0.35\linewidth]{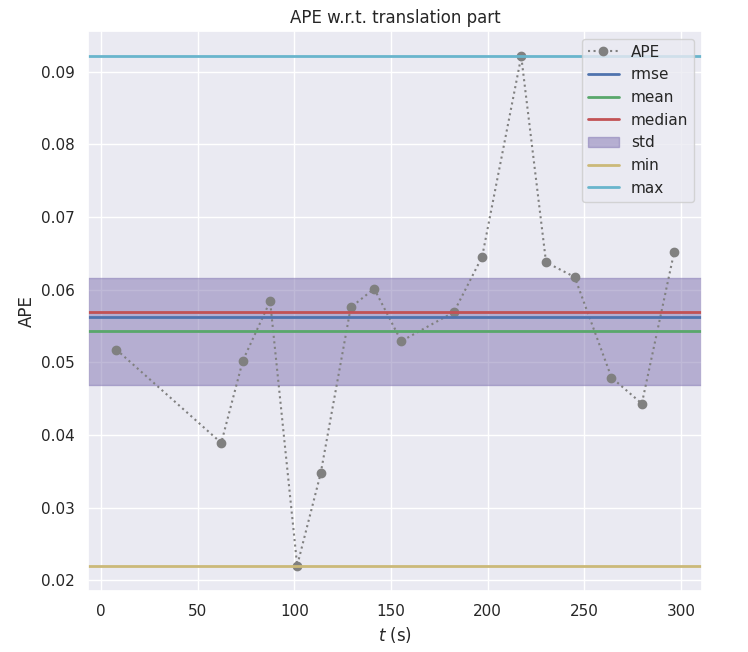}
  \caption{Comparison of the APE (average percentage error) on poses estimated with our SLAM system described in~\secref{sub:expsetup} as reported by HILTI challenge evaluation system on the \texttt{BASEMENT\_3} dataset. This dataset contains scans from Ouster OS0 64 beam LiDAR, which we subsample to 32 beams to increase data sparsity. Left image shows the APE computed from the poses using the baseline normals, while the right image shows the same APE computed from the poses estimated using our proposed method for normal estimation. Using our robust normals in our SLAM system that relies on normal estimation for point cloud alignment improves the quality of the resulting poses as can be seen by comparing the position of the lines. All the lines are lower in terms of APE when using our method for normal estimation (right image).}
  \label{fig:hilti}
\end{figure*}

\begin{figure}[t]
  \centering
  \includegraphics[width=0.99\linewidth]{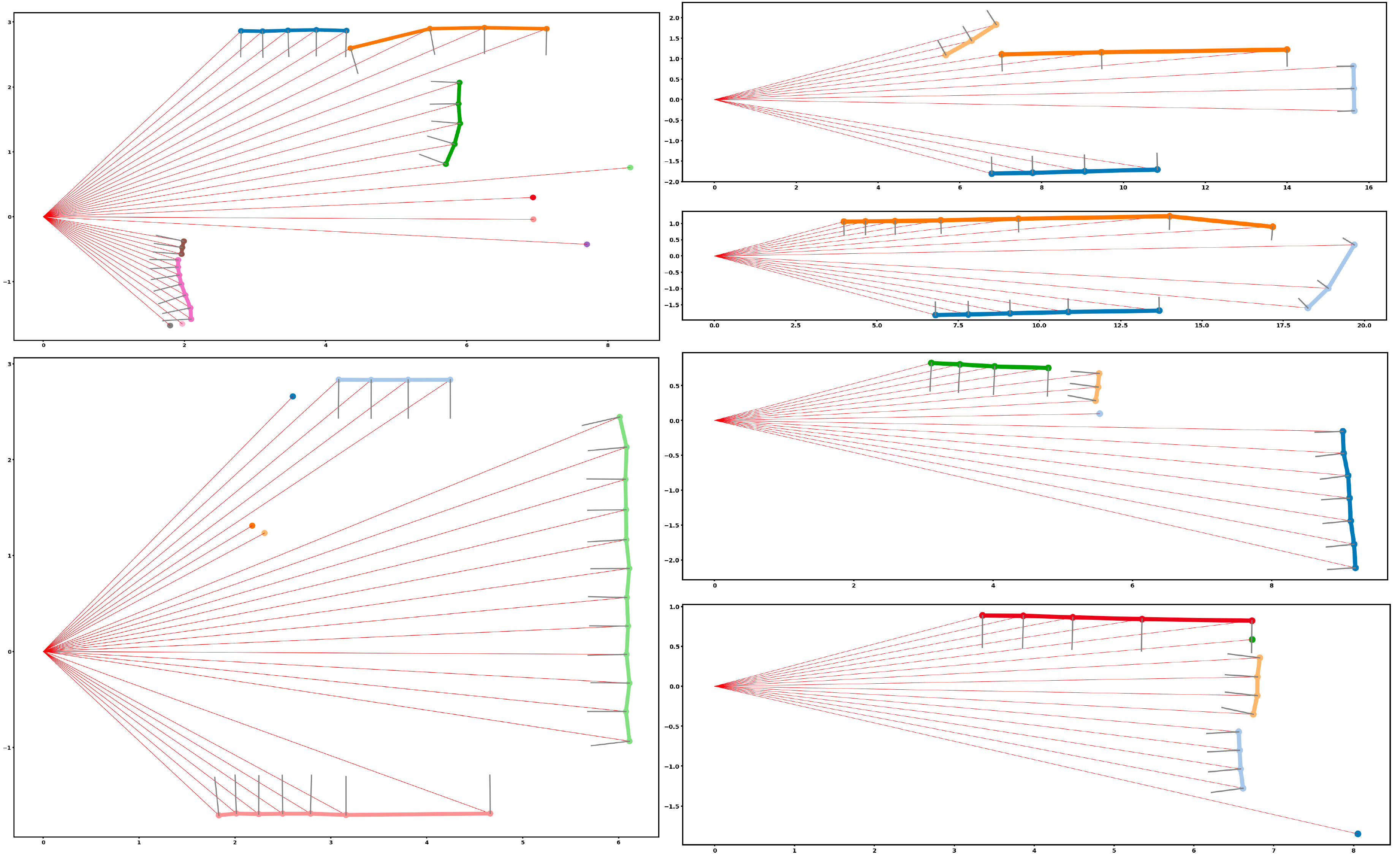}
  \caption{A qualitative assessment of normals computed from 2D slices of 3D data. Every slice is generated from points produced by a single firing of the sensor array of lasers, similar to the illustration in~\figref{fig:normals}. Left side shows 32 beams sampled from a 64 beam Ouster OS0 LiDAR, while the right side shows scans from a 16 beam Velodyne VLP16 LiDAR. The colors of the lines indicate their connected component label, gray lines show the normals estimated at points and the thin red lines show laser beams that generated the points of the scan. Note that some points are missing as the corresponding laser did not return. These results clearly show that our method is able to separate points on different underlying surfaces keeping the normals at fringe points sharp. Note also that standalone points do not have a normal.}
  \label{fig:scans}
\end{figure}

Our SLAM system acquires data from a LiDAR sensor, reconstructs a point cloud, performs motion correction if necessary and works with such point clouds from that point on. As a first step, this point cloud is integrated into a LiDAR odometry system that produces a relative transformation with respect to the previous data. This transformation is then used to add this point cloud to the current submap represented as a voxel grid. Once such a submap has reached a certain size it gets post-processed and added to a node with this submap is added to a pose graph. The submap after the post-processing does not change. We also perform loop closures on the pose graph.

We omit many details here as we believe that they are not fully relevant for this work. What \emph{is} important is that we estimate normals exclusively with the proposed method on single point clouds as soon as they get reconstructed. When these point clouds are processed with our LiDAR odometry system or are integrated into a submap, the normals of all the points that fall into a single voxel get averaged. Therefore, all the dense point clouds that we show in this section indicate the robustness of the normals as detected by our method.

\subsection{Normal Robustness}

The first experiment evaluates the performance of our approach and its outcome supports the claim that the proposed method of computing normals outperforms the baseline in terms of robustness of the produced normals. We evaluate the robustness of the normals in three key ways.

First, we assess the maps built with our SLAM system outlined in~\secref{sub:expsetup} visually. \figref{fig:robustness} showcases a number of places in various environments. We recorded data at the Magic Leap office with a NavVis M6 system that sports a Velodyne VLP16 LiDAR, positioned horizontally. This LiDAR has 16 vertical laser beams that produce sparse range data. In addition to that we make use of publicly available \texttt{BASEMENT\_3} dataset from HILTI~\cite{hilti}. This dataset uses an Ouster OS0 64 beam LiDAR which produces much denser data. We take every second beam in order to showcase the performance of our method with sparse data. The places shown in~\figref{fig:robustness} depict points colored by their normal. A sharper edge between two surfaces indicates less noise in the normals. The images clearly show that the edge between different surfaces is sharper when using the normals computed with the method presented in this paper and so shows their higher robustness.

Another way to evaluate the robustness of the normals is by utilizing them in the SLAM system. If a SLAM system uses normals extensively to align various range data then better normals should lead to better alignment. We performed a SLAM run on the \texttt{BASEMENT\_3} dataset from HILTI~\cite{hilti} using the standard normals as well as the proposed method. The results in~\figref{fig:hilti} show that the error incurred by our SLAM system when using the proposed normal estimation method is lower than the one using the standard method, which proves that the normals generated with the proposed algorithm are more robust.

Last but not least, in~\figref{fig:scans} we show slices of data produced by a single array of vertical lasers from an Ouster OS0 64 LiDAR reduced to 32 beams (left side) and from Velodyne VLP16 LiDAR (right side). These slices are colored by the connected component to which the points and lines correspond and show the computed normals with gray lines. It is easy to see that the proposed algorithm is able to detect normals both inside the "strong" components as well as on their borders.

\subsection{Runtime}

\begin{table}[t]
  \caption{Average runtime and std.~dev.}
  \centering
  \begin{tabular}{C{3cm}cc}
    \toprule
    Method           & 16 beams                & 32 beams                \\
    \midrule
    Our normals      & 1.02\,ms~$\pm$~0.14\,ms & 2.54\,ms~$\pm$~0.61\,ms \\
    Baseline normals & 0.56\,ms~$\pm$~0.02\,ms & 1.3\,ms~$\pm$~0.26\,ms  \\
    \bottomrule
  \end{tabular}
  \label{tab:timing}
\end{table}

Finally we analyze the runtime incurred by our method in comparison to the baseline method. While computing the normals with the proposed method is slower as more actions need to take place, we believe that this cost is warranted in order to achieve higher normals robustness. We report timing for both 16 beam as well as 32 beam LiDARs to show that our method only incurs a constant factor of around 2 as it does not change the \bigO{n} complexity of the baseline method that computes normals. The results of this comparison can be found in~\tabref{tab:timing} that shows the time it takes to compute normals of a single point cloud representing as a single revolution of either a 16 beam or a 32 beam sensor on an Intel\textregistered~Xeon\textregistered~2.10GHz CPU.

\section{Conclusion}
\label{sec:conclusion}

In this paper, we present a novel approach to compute normals on sparse LiDAR data.
Our method exploits the organized structure of such data as well as an assumption that most of the surfaces in the environment are smooth. Utilizing this assumption and separating the points based on the change in the angle of the line segments connecting them allowed us to improve the robustness of the estimated normals while still maintaining competitive runtime of the method.
We implemented and evaluated our approach on various datasets both qualitatively and quantitatively. The experiments suggest that our method outperforms the baseline method in terms of resulting normals robustness while incurring only a constant factor runtime overhead.

\section*{Acknowledgments}
We thank Olga Vysotska for fruitful discussions on the topic of normal estimation and Cyrill Stachniss for his valuable comments.

\bibliographystyle{plain_abbrv}

\bibliography{glorified,new}

\end{document}